\pdfoutput=1

\documentclass[11pt]{article}

\usepackage{authblk}

\usepackage[]{EMNLP2022}

\usepackage{times}
\usepackage{latexsym}
\usepackage{amsmath}
\usepackage{amssymb}
\usepackage{graphicx}
\usepackage{subcaption}
\usepackage{multirow}
\usepackage{changepage}
\usepackage{hhline}
\usepackage{color}
\usepackage{booktabs}
\usepackage{bbm}
\usepackage{wrapfig,lipsum,booktabs}
\usepackage{makecell}
\usepackage{float}
\usepackage{arydshln}

\usepackage[T1]{fontenc}

\usepackage[utf8]{inputenc}

\usepackage{microtype}

\usepackage{inconsolata}

\DeclareMathOperator*{\argmax}{arg\,max}

\title{Tracing Semantic Variation in Slang}

\author[1]{Zhewei Sun}
\author[1,2]{Yang Xu}
\affil[1]{Department of Computer Science, University of Toronto, Toronto, Canada}
\affil[2]{Cognitive Science Program, University of Toronto, Toronto, Canada}
\affil[ ]{\ttfamily \{zheweisun, yangxu\}@cs.toronto.edu}

\begin{document}
\maketitle
\begin{abstract}


The meaning of a slang term can vary in different communities. However, slang semantic variation is not well understood and under-explored in the natural language processing of slang. One existing view argues that slang semantic variation is driven by culture-dependent communicative needs. An alternative view focuses on slang's social functions suggesting that the desire to foster semantic distinction may have led to the historical emergence of community-specific slang senses. We explore these theories using computational models and test them against historical slang dictionary entries, with a focus on characterizing regularity in the geographical variation of slang usages attested in the US and the UK over the past two centuries. We show that our models are able to predict the regional identity of emerging slang word meanings from historical slang records. We offer empirical evidence that both communicative need and semantic distinction play a role in the variation of slang meaning yet their relative importance fluctuates over the course of history. Our work offers an opportunity for incorporating historical cultural elements into the natural language processing of slang.


\end{abstract}

\section{Introduction}

\begin{figure}[t!]
	\begin{subfigure}[b]{0.99\linewidth}
		\includegraphics[width=\linewidth]{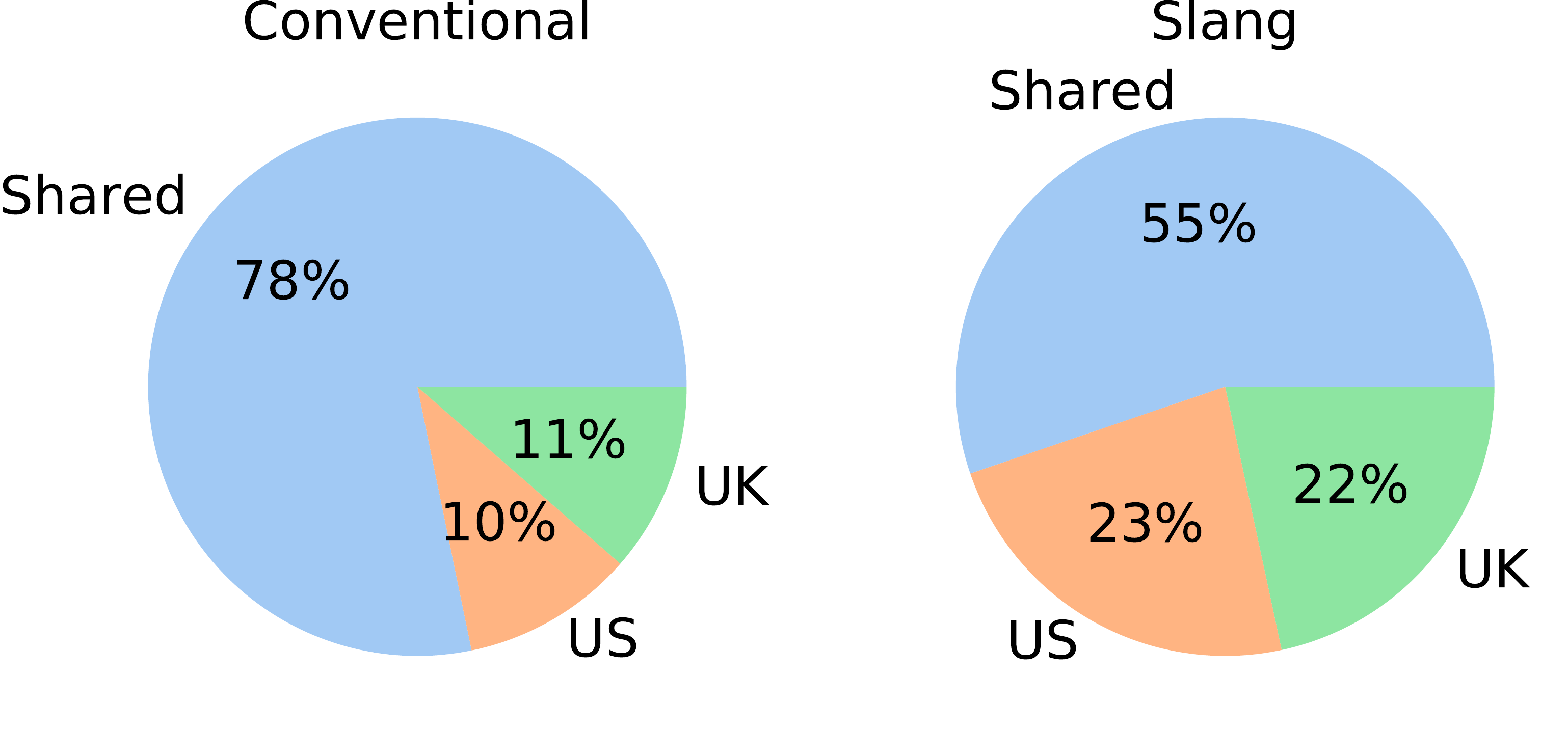}
	\end{subfigure}\vfill
	\caption{Distribution of regional identities among sense entries found in the English Wiktionary. See Appendix~\ref{appwiki} for the detailed experimental setup.}
	\label{figwiki}
\end{figure}

Slang is a type of informal language  commonly used in both day-to-day conversations and online written text. The pervasiveness of slang has generated increasing interest in the natural language processing (NLP) community, with systems proposed for automatic detection~\cite{dhuliawala16, pei19}, generation~\cite{sun19, sun21}, and interpretation~\cite{ni17, sun22} of slang. However, these existing approaches do not account for the semantic variation of slang among different groups of users---a defining characteristic of slang which distinguishes it from conventional language~\cite{andersson92, mattiello05, eble12}. 
Figure~\ref{figwiki} shows a tally of Wiktionary entries confirming that semantic variation is much more prevalent in slang compared to conventional language. Here we develop a principled computational approach to investigate regularity in the semantic variation of slang.

\begin{figure}[t!]
	\begin{subfigure}[b]{0.99\linewidth}
		\includegraphics[width=\linewidth]{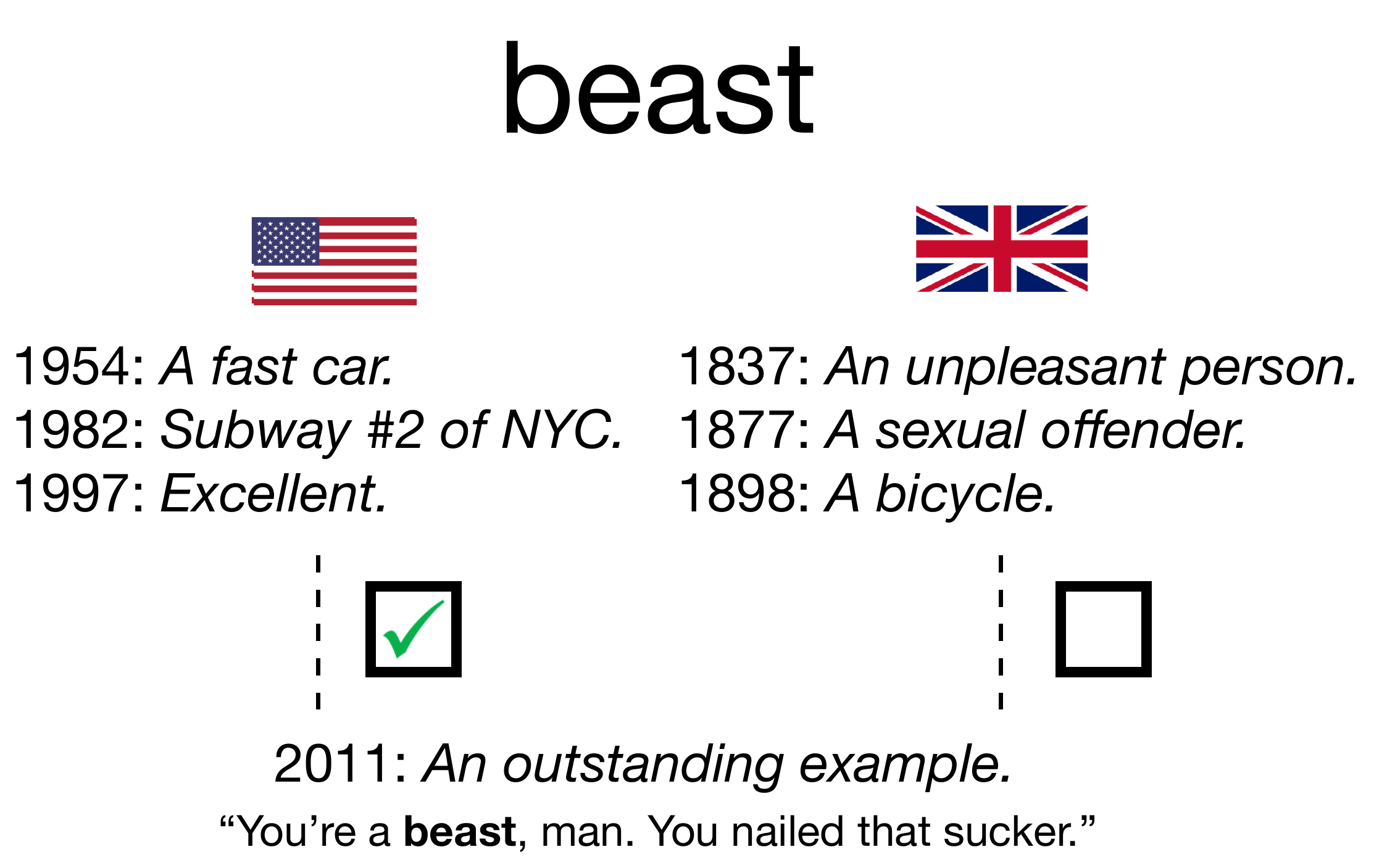}
	\end{subfigure}\vfill
	\caption{Illustration of semantic variation in  the slang word {\it beast}, with senses recorded in American and British English respectively. We develop slang semantic variation models  to trace the regional identity of a new emerging slang sense given its historical meanings and usages from different regions.} 
	\label{fig1}
\end{figure}

We define semantic variation in slang as how a  slang term might take on different meanings in different communities. For example, Figure~\ref{fig1} shows how the commonly used slang word {\it beast} has divergent meanings in different regions (or more specifically, two different countries in this case). Whereas it is often used to express positive things or sentiment in the US, the same slang word has been used to express more negative senses in the UK.



Recent work has quantified  semantic variation in non-standard language of online communities using word and sense embedding models and discovered that community characteristics (e.g., community size, network density) are relevant factors in predicting the strength of this variation~\cite{tredici17, lucy21}. However, it is not clear \textit{how} slang senses vary among different communities and what might be the driving forces behind this variation.

As an initial step to model semantic variation in slang, we focus on regional semantic variation between the US and the UK by considering a regional inference task illustrated in Figure~\ref{fig1}: Given an emerging slang sense (e.g., ``An outstanding example'') for a slang word (e.g., {\it beast}), infer which region (e.g., US vs. UK) it might have originated from based on its historical meanings and usages. Our  premise is that a model capturing the basic principles of slang semantic variation should be able to trace or infer the regional identities of emerging slang meanings over time.




\section{Theoretical Hypotheses}

We consider two theoretical hypotheses for characterizing regularity in slang semantic variation: communicative need and semantic distinction.

{\bf Communicative need.} Prior  work has suggested that  slang  may be driven by culture-dependent \textit{communicative need}~\cite{sornig81}. 
We refer to communicative need as how frequently a meaning needs to be communicated or expressed. Following recent work (e.g.,~\citealt{temp12, ryskina20}), we estimate communicative need based on usage frequencies from Google Ngram\footnote{\url{https://books.google.com/ngrams}} over the past two centuries.\footnote{We acknowledge that experiment-based methods for estimating need exist (see \citealp{karjus21}), but these alternative methods are difficult to operationalize at  scale and in naturalistic settings required for our analysis.} In the context of slang semantic variation, certain things might be more frequently talked about in one region (or country) over another. As such, we might expect these differential needs to drive meaning differentiation in slang terms. For example, a US-specific slang sense for {\it beast} describes the subway line \#2 of the New York City transit network, most likely due to the specific need for communicating that information in the US (as opposed to the UK). 

%

{\bf Semantic distinction.} We also consider an alternative hypothesis termed \textit{semantic distinction} motivated by the social functions  of slang (c.f., \citealp{labov72, hovy90})---language that is used to show and reinforce group identity~\cite{eble12}. Under this view, slang senses may develop independently in each region and form a semantically cohesive set of meanings that reflect the cultural identity of a region. As a result, emerging slang senses are more likely to be in close semantic proximity with historical slang senses from the same region.\footnote{It is worth nothing that communicative need and semantic distinction may not be completely orthogonal. In fact, differences in communicative need may drive semantic distinction. However, we consider these hypotheses as alternative ones because they are motivated by different functions.} For example, the slang {\it beast} has formed a cluster of senses in the US that describes something virtuous while senses in the UK often describe criminals. An emerging sense such as ``An outstanding example'' would be considered more likely to originate from the US due to its similarity with the historical US senses of {\it beast}. Here we operationalize semantic distinction by models of semantic chaining from work on historical word meaning extension~\cite{ramiro18, habibi20}, where each region develops a distinct chain of related regional senses over history.

We evaluate these theories using slang sense entries from Green's Dictionary of Slang (GDoS, \citealp{green10}) over the past two centuries. Analysis on GDoS entries is appropriate because 1) a more diverse set of topics is covered compared to domain-specific slang found in online communities (e.g., Reddit), and 2) the region and time metadata associated with individual sense entries support a diachronic analysis on slang semantic variation.
To preview our results, we show that both communicative need and semantic distinction are relevant factors in predicting slang semantic variation, with an exemplar-based chaining model offering the most robust results overall.
Meanwhile, the relative importance of the two  factors is time-dependent and fluctuates over different periods of history.

\section{Related Work}

\subsection{Variation in online language}

Previous  work in computational social science on online social media has explored lexical variation~\cite{eisenstein14, nguyen16} by studying the differences in word choice among different online communities. It has also been shown that linguistic and social variables can predict the popularity and dissemination of linguistic innovations in online language~\cite{stewart18, tredici18}. \citet{yang17} modeled sentiment variation of words found in tweets, where users with close ties are assumed to give similar sentiment labels for the same word. \citet{tredici17} adapted \citet{bamman14}'s distributive embedding model to train community-specific word embeddings for a small set of Reddit communities and quantified semantic variation by comparing cosine similarities between community-specific embeddings for the same word.

\citet{lucy21} extended the previous study to quantify semantic variation of online language in 474 reddit communities. They compared PMI based sense specificity of clustered BERT~\cite{devlin19} embeddings generated using different contextual instances of a word's usage, along with an alternative strategy that uses BERT to predict word substitutions from the same usage instances~\cite{amrami19}. \citet{lucy21} also proposed a regression-based model of semantic variation with community-based features (e.g., community size, network density) as well as topical features derived from Reddit's subreddit hierarchy. While they find these features to be informative in predicting the strength of semantic variation, they do not explicitly model how slang senses vary. Instead of predicting the strength of semantic variation, our work takes a more direct approach by modeling how slang senses vary and study the driving forces underlying such variation. We also extend our analysis to study attested slang usages over the past two centuries instead of focusing on contemporary internet slang.

Also related to our work is \citet{keidar22} who performed a causal analysis of slang semantic change using tweets from 2010 to 2020. Slang's usage frequencies were found to change more drastically than those of conventional language while the semantic change for stable senses progresses much slower. In our study, we make a complementary observation in which slang senses from the 19th century are still relevant for predicting semantic variation in contemporary slang.







\subsection{NLP for slang}

Recent work in natural language processing has shown increasing interest in the automatic processing of novel slang, moving beyond retrieval based methods~\cite{dhuliawala16, wu18, gupta19} that do not generalize to emerging slang usages absent in training.
In particular, end-to-end deep neural networks have been proposed for slang detection~\cite{pei19}, slang interpretation~\cite{ni17}, as well as the formation of slang words~\cite{kulkarni18}.
\citet{sun21} proposed a model of slang semantics based on Siamese networks~\cite{Baldi93, Bromley94} to learn joint representations for both conventional and slang senses. The resulting sense representations can then be used with a semantic chaining model~\cite{ramiro18} to generate novel slang usages~\cite{sun19, sun21} or better interpret slang usages in text~\cite{sun22}. In those cases, each candidate word $w$ is considered a class and its conventional senses are taken as class attributes of $w$. We apply chaining models in different ways from those in \citet{sun21}. Instead of treating each word as a class, we group senses by region and consider only slang senses. 

Previous NLP approaches to slang have often assumed that slang expressions are homogeneous across different groups of users. Here, we relax this assumption by explicitly modeling the factors that contribute to slang semantic variation. 
For example, a slang interpreter could benefit from a semantic variation model in cases where the  region has been pre-determined, so that an interpreter would prefer the meaning ``excellent'' when interpreting the slang {\it beast} if the slang is known to be used in the US. We hope that our work will contribute to more sophisticated approaches toward the modeling of informal language for these downstream tasks and real-world applications.



\begin{figure*}[t!]
	\begin{subfigure}[b]{0.33\linewidth}
		\caption{Senses}
		\includegraphics[width=\linewidth]{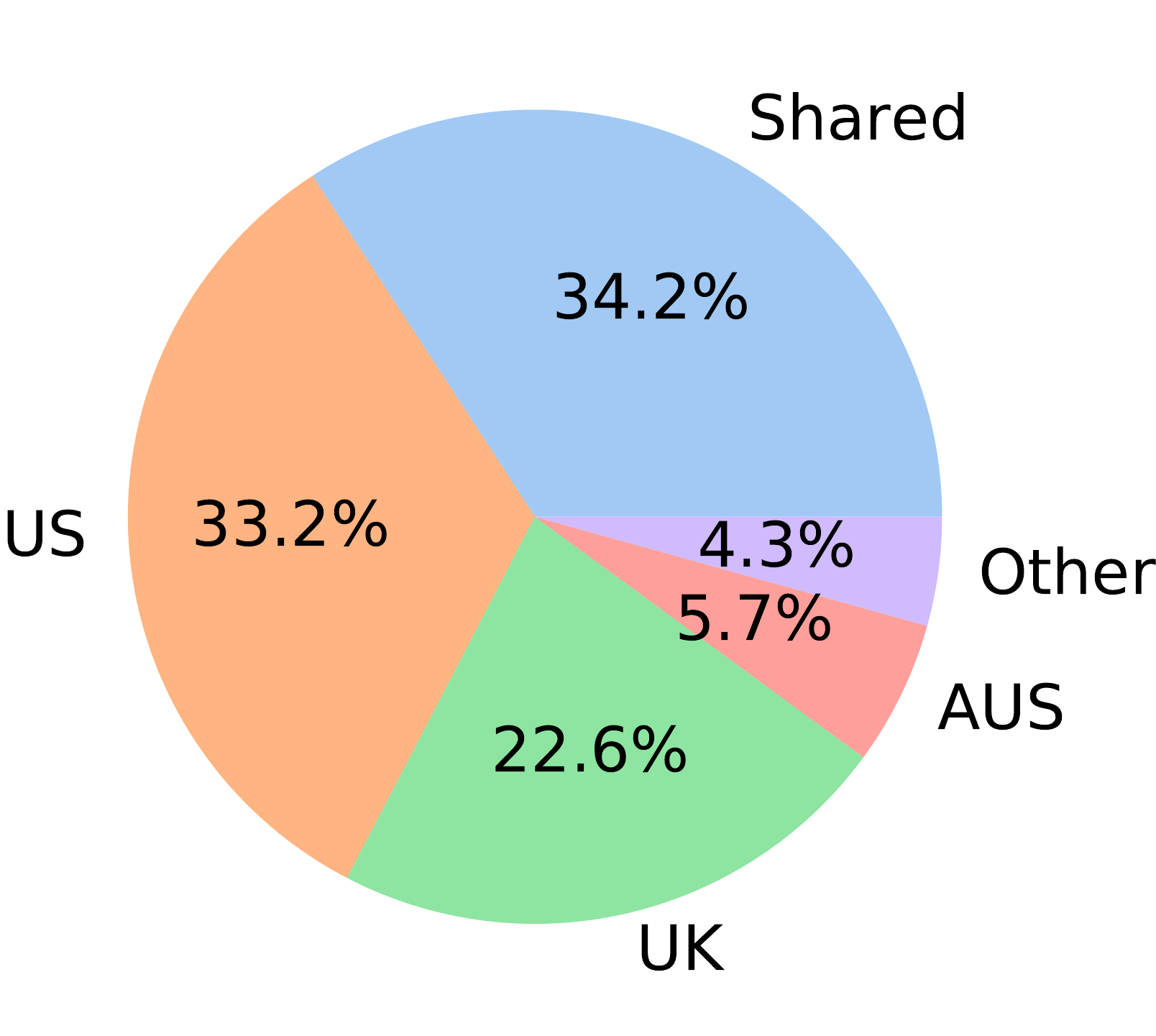}
		\label{figdata_a}
	\end{subfigure}
	\begin{subfigure}[b]{0.33\linewidth}
		\caption{Word forms}
		\includegraphics[width=\linewidth]{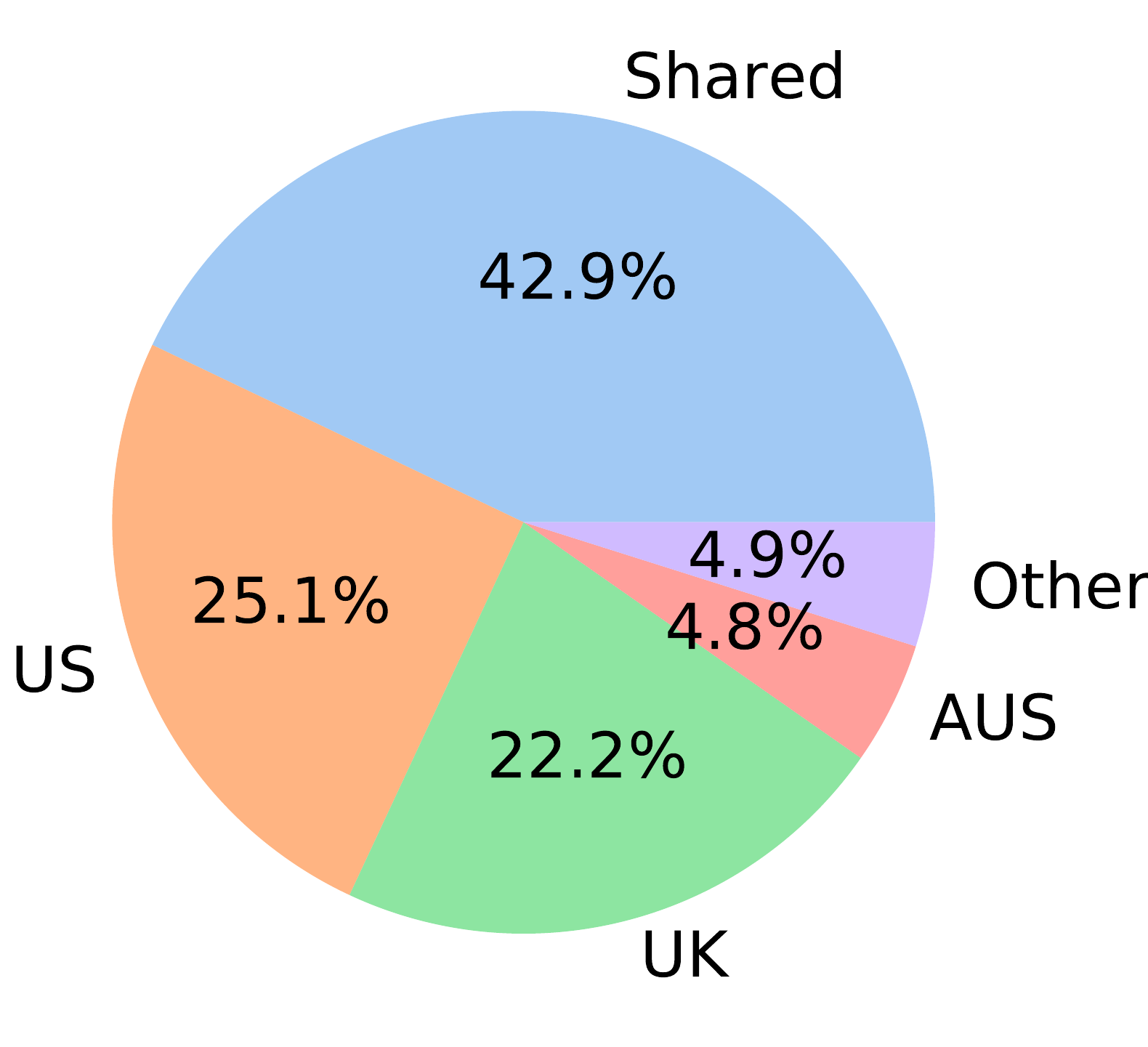}
		\label{figdata_b}
	\end{subfigure}
	\begin{subfigure}[b]{0.33\linewidth}
		\caption{Senses of shared word forms}
		\includegraphics[width=\linewidth]{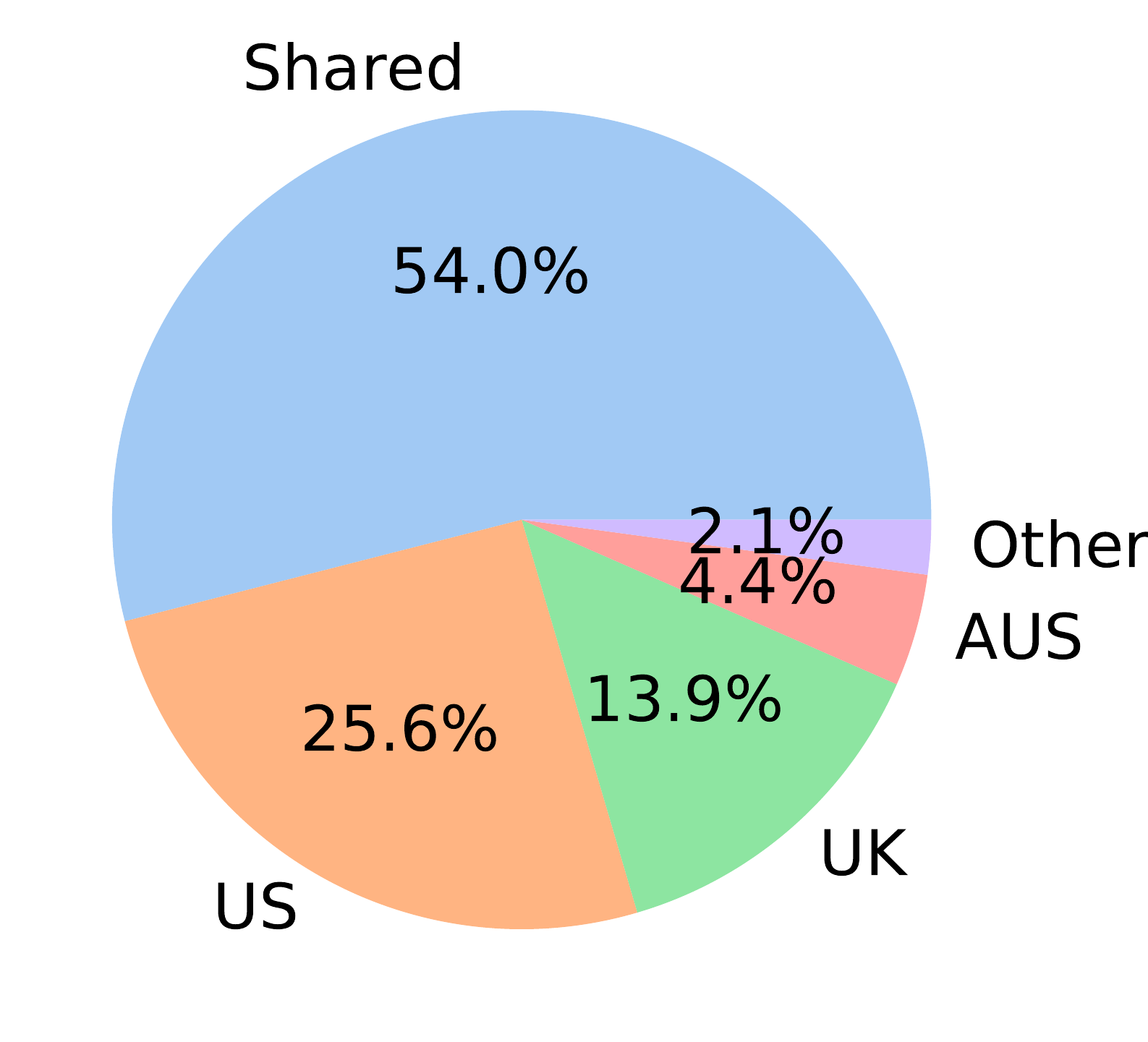}
		\label{figdata_c}
	\end{subfigure}
	\caption{The distribution of GDoS slang senses and word forms across different regions. A word or sense is considered shared if two or more distinct region tags can be found in the constituent references.} 
	\label{figdata}
\end{figure*}

\section{Data}

\subsection{Green's Dictionary of Slang (GDoS)}

We collect slang lexical entries from Green's Dictionary of Slang (GDoS, \citealp{green10})\footnote{\url{https://greensdictofslang.com/}}, a historical English slang dictionary covering more than two centuries of slang usage. Each word entry (e.g., ``beast'') in GDoS is associated with one or more sense entries. 
A sense entry contains a definition sentence (e.g., ``An outstanding example.'') and a series of references. Each reference contains a region tag (e.g., US or UK), a date tag (e.g., 2011), and a sentence indicating the origin of the reference. In some cases, the reference contains an example usage sentence of how the slang is used in context.\footnote{We choose GDoS over alternative resources (e.g., \citealp{lewin88, dalzell09, ayto10}) because it covers a diverse set of slang usages from different regions and time periods.}

We collect all sense entries with at least one valid reference. A reference is considered valid if both its region tag and date tag are not missing nor invalid. For each reference, we automatically extract the associated context sentence and consider one to be valid if it contains precisely one exact occurrence of the word in the sentence. If a valid context sentence is found then it is attached to the corresponding reference. The resulting sense entry may have none or more than one context sentences. In the latter case, we select the context sentence with the earliest time tag to be associated with the sense entry, so that it best represents the usage context of when the sense first emerges. The earliest time tag found in the references is considered the time of emergence for a sense entry. We filter all abbreviation entries as these entries don't create new meaning. In the case of a homonym (i.e., multiple word entries for the same word form), we collapse all entries into a single word entry. After pre-processing, we obtain 42,758 distinct words with 76,650 associated sense entries. On average, each sense entry contains 4.48 tags of attested time and region. We provide our pre-processing script in our Github repository\footnote{Code and data scripts available at: \url{https://github.com/zhewei-sun/slangsemvar}} to facilitate reproduciblility.


\subsection{Data analysis}


We first analyze entries collected from GDoS to quantify semantic variation. For each sense entry, we determine its regional identity using the region tags associated with each reference. Note that there may be more than one valid region tags associated with each sense entry. In such cases, we consider the sense entry to be a shared sense across all constituent region tags. Otherwise, the sense entry is considered regional. Likewise, a word entry is considered shared if two or more distinct region tags can be found among any of its sense's references.


Figures~\ref{figdata_a} and \ref{figdata_b} show the distribution of region identities across all sense and word entries in GDoS. We observe substantial lexical variation within the data where more than half of the word forms are regional. While most of the sense entries are also regional, many of them may be associated with regional word forms. In this case, the variation is caused by difference in lexical choice and does not entail semantic variation. We control for lexical variation by only considering sense entries associated with shared word forms. This results in 48,565 sense entries with an average of 5.80 tags per entry. Figure~\ref{figdata_c} shows the distribution of the resulting region identities. We observe that even after controlling for lexical variation, roughly half of the senses remain regional. Moreover, much of the semantic variation is captured by the US and UK regions, with Australian slang also making up a notable portion. We therefore focus on modeling  semantic variation between the two most represented regions.

\section{Modeling Semantic Variation}

\subsection{Predictive task}

We model semantic variation by formulating a regional inference task: Given an emerging slang sense $s$ for a word $w$, infer the region $r \in \mathcal{R}$ from which the emerging sense originates. Here, $\mathcal{R}$ is the set of regions being considered and an example would be the set $\{$US,  UK$\}$. A semantic variation model $\mathcal{V}$ is then defined as follows:

\begin{align}
    P(r) \propto \mathcal{V}(s, w, r)
\end{align}
Here, the semantic variation model $\mathcal{V}$ captures the likelihood of observing the emerging slang sense $s$ expressed using word $w$ within region $r$ and can be either generative or discriminative in nature. Given the semantic variation model, the target region can be predicted by maximizing the likelihood:

\begin{align}
    r^* = \argmax_{r \in \mathcal{R}}\mathcal{V}(s, w, r)
\end{align}
An effective semantic variation model should prefer regions that are more likely for the new sense to emerge. We next describe models of semantic variation $\mathcal{V}$ motivated by both communicative need and semantic distinction.

\subsection{Models based on communicative need}

We first describe a set of semantic variation models $\mathcal{V}$ inspired by the communicative need principle. Under this hypothesis, language users in different communities necessitate differing expressions to express concepts of particular interest to the community~\cite{sornig81}. We operationalize communicative need using frequency statistics from historical corpora originated from each region. First, we propose a \textit{form need} model that considers the frequency of the slang word form $w$:

\begin{align}
    \mathcal{V}_{\text{form\_need}}(\cdot) \propto f_{r, s_t-\alpha:s_t}(w)
\end{align}
Here, $f_{r, s_t-\alpha:s_t}(w)$ is the frequency of observing word $w$ from region $r$ within a time window $\alpha$ strictly preceding the sense's time of emergence $s_t$. Note that the form need model does not take into account any semantic information from the emerging sense $s$ and simply estimates whether the word form $w$ is more prevalent in one region. The \textit{semantic need} model incorporates semantic information by checking the frequency of all content words within the definition sentence $s_d$ of sense $s$:

\begin{align}
    \mathcal{V}_{\text{semantic\_need}}(\cdot) \propto \sum_{c \in \text{content}(s_d)} f_{r, s_t-\alpha:s_t}(c)
\end{align}
Similarly, the \textit{context need} model is informed by the usage context sentence $s_c$ of sense $s$:

\begin{align}
    \mathcal{V}_{\text{context\_need}}(\cdot) \propto \sum_{c \in \text{content}(s_c) \backslash w} f_{r, s_t-\alpha:s_t}(c)
\end{align}
We remove the word $w$ since it is not part of the context. The context need model checks the communicative context to estimate contextual relevance with respect to each region. Both of the above models can also be framed as a majority vote model instead of taking a sum of frequencies:

\begin{align}
    \mathcal{V}(\cdot) \propto \sum_{c} \mathbbm{1}_{\max_r f_{r, s_t-\alpha:s_t}(c)}f_{r, s_t-\alpha:s_t}(c)
\end{align}
We find the majority vote scheme to be robust in our experiments as frequency counts of common words could otherwise  dominate the estimates.

\subsection{Models based on semantic distinction}

Slang semantics may also diverge due to its social function, where language users in a community wish to create distinct senses to express their social identity~\cite{eble12}. As a result, slang senses from different communities might evolve into cohesive but distinct clusters. Motivated by this hypothesis, we model semantic variation using historical slang senses associated with the word $w$ in a region $r$ that emerged before $s_t$, denoted as $\mathcal{S}_{w,s_t,r}$. Under this paradigm, the semantic variation model $\mathcal{V}$ can be specified as follows:

\begin{align}
    \mathcal{V}_{\text{distinction}}(\cdot) \propto g(s, \mathcal{S}_{w,s_t,r}) \label{eqsd}
\end{align}
Here, the function $g$ can be viewed as a classifier that measures the categorical similarity between the emerging sense $s$ and historical senses from region $r$. We model $g$ generatively using semantic chaining models from historical word sense extension which are motivated by mechanisms of human categorization ~\cite{rosch75, nosofsky86}.\footnote{Sentential context can be potentially integrated to achieve higher accuracies but here we focus on senses alone to examine the effect of semantic cohesiveness operationalized by cognitively motivated modeling approaches.} We adapt three prominent variants of semantic chaining from \citet{ramiro18}: 1) the \textit{one nearest neighbor (onenn)} model that only considers the most similar historical sense;  2) the mean \textit{exemplar} model that accounts for all historical senses; and 3) the \textit{prototype} model which collapses all historical senses into a single prototypical sense. When performing chaining, each sense is represented by embedding its corresponding definition sentence $s_d$ using a sentence embedder $E$:

\begin{align}
    g_{\text{onenn}}(\cdot) & = \max_{s'\in \mathcal{S}_{w,s_t,r}} sim(E(s_d), E(s'_d))\\
    g_{\text{exemplar}}(\cdot) & = \overline{\sum_{s' \in \mathcal{S}_{w,s_t,r}}sim(E(s_d), E(s'_d))}\\
    g_{\text{prototype}}(\cdot) & = sim\Bigg(E(s_d), \overline{\sum_{s' \in \mathcal{S}_{w,s_t,r}} E(s'_d)}\Bigg)
\end{align}
The similarity between two sense embeddings is computed using negative exponentiated distance with a learnable kernel width parameter $h$:

\begin{align}
    sim(e, e') = \exp\Big(-\frac{||e - e'||_2^2}{h}\Big) \label{eqkernel}
\end{align}
When data is available, the kernel width parameter $h$ can be optimized by constructing training examples from the full set of historical senses $\mathcal{S}_{w,s_t}$.



\section{Experiments}

\subsection{Setup}



We test our semantic variation models on region inference using GDoS word entries that show high regional variation in their senses. Specifically, we consider all word entries with at least $k$ regional senses that have emerged after 1800 in each region of interest. We consider $k \in [3, 10]$ and Table~\ref{tablek} shows the number of words and senses that match the criteria for each $k$ when considering \{US, UK\} as the set of regions.
All senses emerged after 1900 are treated as a time series of test examples. For example, all senses of a word that emerged before 1900 will be used as historical senses when predicting the region for the first sense post 1900 and this sense will then be considered as a historical sense when making a prediction for the subsequent sense. 
Word entries with sparse regional senses (i.e., $k=1$ or $k=2$) are excluded because they often result in uninformative test examples where not a single slang usage in one region is available prior to 1900.

\begin{table}[t!]
	\centering\makebox[0.5\textwidth]{
		\begin{tabular}{lrrrrr}
			k&\makecell[r]{Word  \\entries}&\makecell[r]{US  \\senses}&\makecell[r]{UK  \\senses}&\makecell[r]{Shared  \\senses}&\makecell[r]{Test  \\set}
			\\
			\addlinespace[0.05cm]
			\hline
			\addlinespace[0.1cm]
			3 &388&3273&1889&2007&1722\\
			\addlinespace[0.05cm]
			4 &209&2063&1263&1272&1200\\
			\addlinespace[0.05cm]
			5 &114&1342&827&877&788\\
			\addlinespace[0.05cm]
			6 &64&842&550&577&548\\
			\addlinespace[0.05cm]
			7 &44&627&423&446&424\\
			\addlinespace[0.05cm]
			8 &30&455&316&337&310\\
			\addlinespace[0.05cm]
			9 &21&286&239&240&230\\
			\addlinespace[0.05cm]
			10 &14&192&176&156&162\\
	\end{tabular}}
	\caption{Number of GDoS word and sense entries obtained after constraining the minimum number of regional senses per region ($k$). Senses are divided into regional and shared based on region tags associated with sense references. The last column shows the sizes of the test sets where each is comprised of an equal number of test senses from each region.}
	\label{tablek}
\end{table}

Since there are often a disproportionate number of test senses between the two regions, we create class-balanced test samples by subsampling eligible test senses in each time series. For example, a word entry with 5 US sense entries and 3 UK sense entries emerged after 1900 will result in 6 test examples where 3 out of the 5 US senses are randomly sampled while all UK senses are kept. Even if a sense entry has not been sampled for prediction, it will still appear in the history when predicting subsequently emerged senses. The last column of Table~\ref{tablek} shows the sizes of the class-balanced test samples where half of the sense entries come from each region. To account for all senses in the data, we repeat the sampling procedure 20 times in all of our experiments and report the mean predictive accuracy. Word lists for each $k$ can be found in our Github repository.

We use case-insensitive normalized frequency from the 2019 version of Google Ngram's ``American English'' and ``British English'' corpora to estimate word frequencies for all communicative need models and set the window size $\alpha$ to 10 years. The list of stopwords from NLTK~\citep{bird09} is used to filter for content words. We apply additive smoothing of $1e^{-8}$ and 1 to normalized frequency and majority vote models respectively. In the case of a tie, the model defaults to predicting US.

\begin{table*}[t!]
	\centering\makebox[\textwidth]{
		\begin{tabular}{lrrrrrr}
		    
		    &\multicolumn{3}{c}{No shared senses}&\multicolumn{3}{c}{With shared senses}
			\\
			Model&US senses&UK senses&All senses&US senses&UK senses&All senses
			\\
			\addlinespace[0.05cm]
			\hline
			
			\addlinespace[0.1cm]
			Form need & 49.9 (1.20) & 50.7 (0.20) & 50.3 (0.57)\\
			Semantic need & 54.0 (1.77) & 50.1 (0.43) & 52.1 (0.91) & \multicolumn{3}{c}{Not applicable}\\
			Context need & 65.9 (1.37) & 43.8 (0.46) & \textbf{54.8} (0.67)\\

			\addlinespace[0.15cm]
			Sense frequency & 55.9 (1.44) & 34.1 (0.26) & 45.0 (0.75) & 59.2 (1.33) & 34.0 (0.27) & 46.6 (0.66)\\
			LDA & 51.7 (1.37) & 45.3 (0.35) & 48.5 (0.73) & 54.7 (1.46) & 45.5 (0.50) & 50.1 (0.81)\\
			Logistic reg. & 52.5 (1.57) & 40.0 (0.35) & 46.2 (0.84) & 56.5 (1.24) & 39.3 (0.32) & 47.9 (0.64)\\
			
			\addlinespace[0.15cm]
			Onenn & 60.9 (1.35) & 53.0 (0.42) & 56.9 (0.71) & 72.4 (1.32) & 38.0 (0.41) & 55.2 (0.64)\\
			Exemplar & 60.1 (1.31) & 57.8 (0.40) & \textbf{58.9} (0.70) & 60.0 (1.66) & 58.6 (0.38) & \textbf{59.3} (0.85)\\
			Prototype & 57.6 (1.38) & 53.1 (0.37) & 55.4 (0.77) & 60.3 (1.71) & 54.7 (0.30) & 57.5 (0.88)\\
	\end{tabular}}
	\caption{Mean percentage accuracy of all models on the region tracing task for post 1900 senses associated with words that have at least 5 regional senses in each region (US and UK; $k=5$). Standard deviation of accuracies taken across 20 test samples is shown in parenthesis. The right-hand side shows the results after including shared senses in both training and prediction.}
	\label{tablemain}
\end{table*}

For semantic distinction based models, sense embeddings are obtained by embedding their respective definition sentences from GDoS using Sentence-BERT (SBERT, \citealp{reimers19}). In addition to the semantic chaining models, we include \textit{LDA} and \textit{logistic regression} as discriminative baselines for the classifier $g$ in Equation~\ref{eqsd} where each sense's definition sentence is encoded using SBERT and used as the feature vector. We also include a \textit{sense frequency} baseline that always predicts the most frequent sense tag observed in the historical senses. When all historical senses correspond to a single region label, then that label is taken as the prediction for all semantic distinction models.


To train the kernel width parameter in semantic chaining (i.e., $h$ in Equation~\ref{eqkernel}), we consider all historical senses as a time series and train on as many predictions as data allows. For example, if a list of senses emerged prior to the to-be-predicted sense, then we iterate through these senses in their order of emergence. As soon as there is an observation from each class, chaining probabilities are estimated and the negative log-likelihood of the corresponding region is included in the loss function for optimization. Following \citet{sun21}, we use L-BFGS-B~\citep{byrd95} to optimize the kernel width with a default value of 1 and a bound between [$0.01$, $100$].

\subsection{Inference of  regional identity of slang} \label{secmainexp}

We now evaluate both the communicative need and semantic distinction based semantic variation models on the regional inference task. Table~\ref{tablemain} shows the mean predictive accuracy of all models on the $k=5$ test set. For communicative need, we observe that both the semantic need and context need models made better predictions than the simple form need model and the random baseline (i.e., 50\% accuracy). For semantic distinction, the chaining models consistently outperform both the random baseline and the discriminative classifiers.\footnote{We also considered a simple hybrid model that combines likelihood estimates from both communicative need and semantic distinction models but observe no better result.Results for this experiment can be found in Appendix~\ref{appcombo}} 
While the standard classifiers' (LDA and logistic regression) predictability suffers from data sparsity, both models of slang variation are able to leverage enriched data points from history to perform well in a few-shot setting.
Indicated by poor performance from the sense frequency baseline, we observe no discernible patterns in the emergence trajectory of senses across regions when the content of the slang usage is disregarded.
In line with previous applications of semantic chaining to linguistic categories~\cite{habibi20, yu21}, we also find exemplar-based chaining performing the best among alternative chaining models, suggesting that regional slang senses tend to form cohesive neighborhoods in the underlying semantic space. 
Both the exemplar and prototype models also tend to rely less on sense frequency and produce more balanced predictive accuracies across the two regions. We observe similar trends in a 3-way classification experiment involving Australian slang shown in Appendix~\ref{appaus}.

We also consider the inclusion of shared senses in addition to their regional counterparts. As shown in Table~\ref{tablek}, shared senses account for a large portion of the sense inventory that could result in more data. For each shared sense, we track its list of references to determine its regional identity at a particular point in history. For example, a sense containing a US reference in 1930 and a UK reference in 1940 would be considered US exclusive when used to predict the region of a sense emerging between 1930 and 1940. Senses with shared regional identities (e.g., the aforementioned sense after 1940) are considered by both regional categories in semantic chaining models to obtain more accurate kernel width estimates. We observe better model performance in all models that consider historical senses after including shared senses. Introducing shared senses most notably improved the prototype model where more senses arguably led to more accurate estimates of the prototypical senses.

\begin{figure}[t!]
	\begin{subfigure}[b]{0.99\linewidth}
		\includegraphics[width=\linewidth]{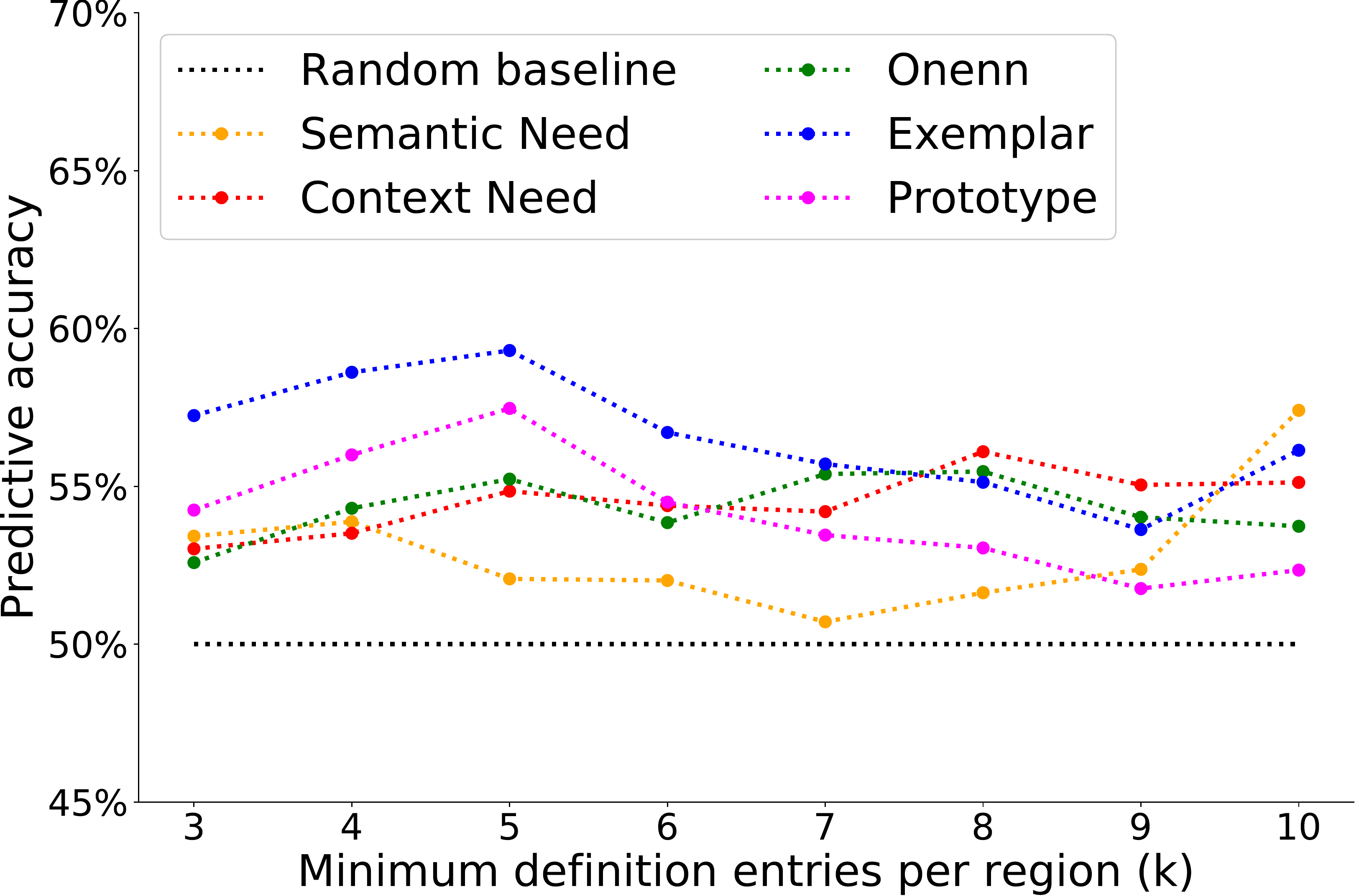}
	\end{subfigure}\vfill
	\caption{Predictive accuracy of the best performing models relative to the minimum number of regional senses ($k$) in sampled word entries. All shared historical senses are used in semantic chaining.} 
	\label{figresk}
\end{figure}

Figure~\ref{figresk} shows the predictive accuracy of the best performing models over all samples of $k$.
Overall, both communicative need and semantic distinction models are able to capture a notable amount of variation in the data, with the semantic chaining based models giving the best predictability.
Also, the advantage of chaining-based models over frequency-based models diminishes for more polysemous slang (which presumably is also more frequently used). This suggests that as a slang obtains more senses, its set of senses becomes less cohesive and the slang word is more likely to be used to express concepts with high communicative need that are more coarsely related to its historical meaning. An alternative explanation is that those historical senses become conventionalized or dismissed over history and are thus no longer relevant in the emergence of new slang senses. Next, we test this hypothesis by constraining the number of senses considered in the chaining models.

\subsection{Memory in semantic variation}

\begin{figure}[t!]
	\begin{subfigure}[b]{0.99\linewidth}
		\includegraphics[width=\linewidth]{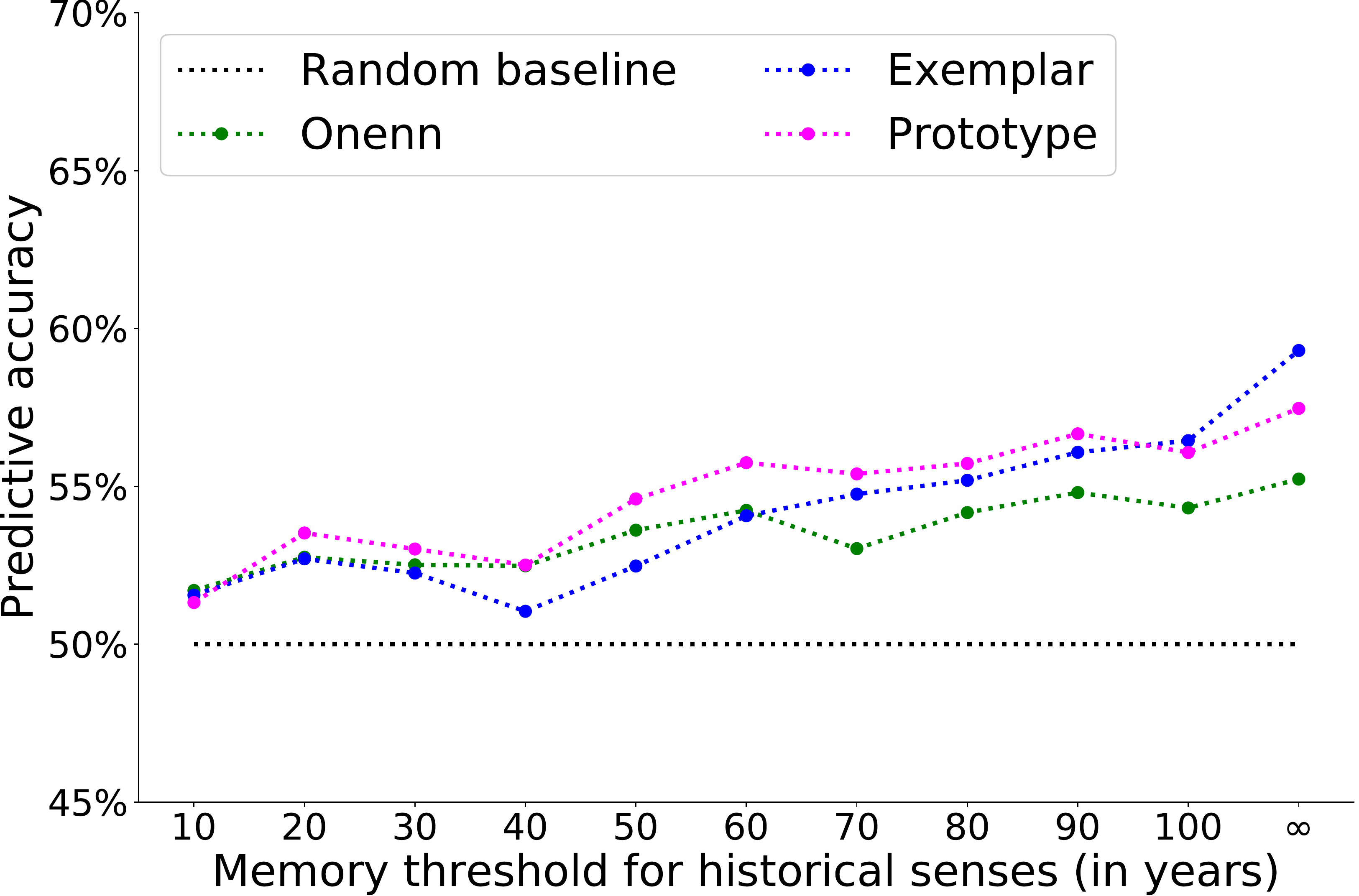}
	\end{subfigure}\vfill
	\caption{Predictive accuracy of all chaining models with shared senses after removing historical senses that exceed the memory threshold during prediction.} 
	\label{figresmem}
\end{figure}

Slang senses are known to be short-lived and become conventionalized or dismissed over time~\cite{eble89}. We measure to what extent historical senses are relevant in the process of variation. We do so by constraining the number of historical senses seen by the chaining models based on their year of emergence. We focus on the $k=5$ case and find that without a memory constraint, the average age of historical senses ranges from 36.8 years for test senses in the 1910s to 73.7 years for those in the 2010s. Figure~\ref{figresmem} shows the mean predictive accuracy for all chaining models after removing historical senses that exceed the memory threshold. To preserve model efficacy, historical senses can still be used as examples to train the kernel width parameter, but those examples themselves are also restricted to historical senses within the memory threshold when making predictions during training. Despite our intuition, we observe a consistent upward trend in predictive accuracy as the memory constraint becomes more relaxed. Historical slang senses dating over 100 years nevertheless remain relevant when considering the semantic variation of contemporary slang.

\subsection{Historical analysis of semantic variation} \label{sechist}

As a final analysis, we examine the relative importance of communicative need and semantic distinction over history by comparing their respective best-performing models on emerging senses over the past millennium. We consider all senses that emerged after 1900 in the $k=5$ condition without class-balance sampling and partition all senses by their decade of emergence. For each decade, we create class-balanced test samples by sampling the set of senses from the more frequent region to match in size. We repeat this sampling procedure 20 times to account for all senses. We show the resulting mean predictive accuracy in Figure~\ref{figreshist} and test sample sizes in Appendix~\ref{apphistc}. Overall, both the communicative need and semantic distinction based models show  fluctuations in performance over different periods of history, More interestingly, we observe several decades in which the two models show polarizing results. In time periods such as the 1940s and 1990s, communicative need becomes the stronger predictor compared to semantic distinction. This may attribute to historical events such as World War II and the advent of internet where language users were exposed to new inventions and concepts that require slang to express, whereas language users may become more motivated in creating distinct cultural identities during the post-war eras.

\begin{figure}[t!]
	\begin{subfigure}[b]{0.99\linewidth}
		\includegraphics[width=\linewidth]{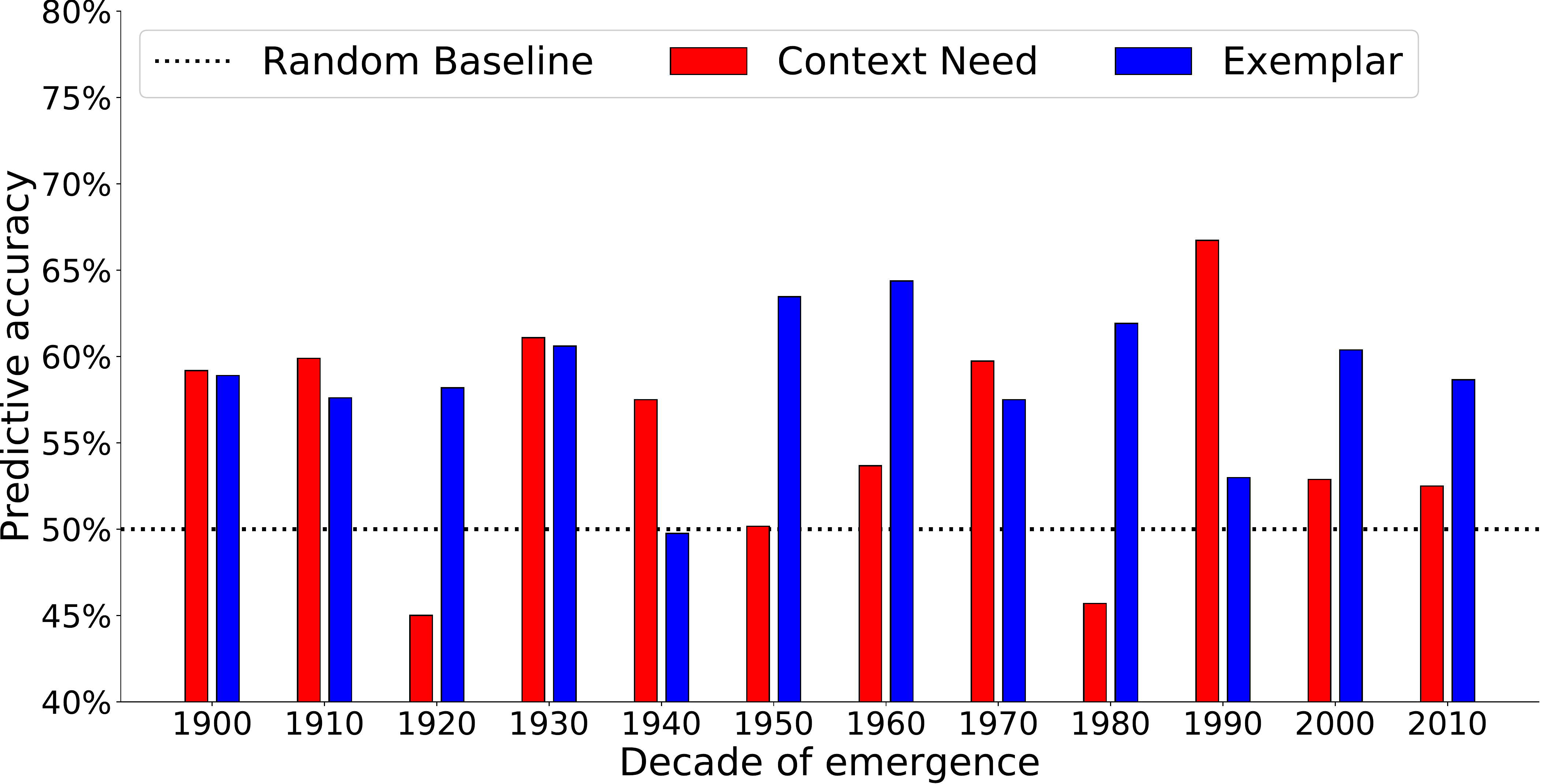}
	\end{subfigure}\vfill
	\caption{Predictive accuracy of the best performing communicative need and semantic distinction based models in different periods of history. All shared historical senses are used in semantic chaining.} 
	\label{figreshist}
\end{figure}

\section{Conclusion}

We have presented a principled and large-scale computational analysis of slang semantic variation over history. Inspired by the theoretical hypotheses of communicative need and semantic distinction, we develop a computational approach to test these theories against regional slang attested in the UK and the US over a stretched period of time. We find regular patterns in slang semantic variation that are predictive of regional identities of the emerging slang senses.
While both hypotheses are found to be relevant for predicting slang semantic variation, we observe that semantic distinction  better explains the semantic  variation in slang terms used in the US and the UK over the past two centuries. Our work sheds light on the basic principles of slang semantic variation and provides opportunities for incorporating historical cultural elements in the automated processing of informal language such as slang generation and interpretation.

\section*{Limitations}
In our work, we have restricted our scope of investigation to regional semantic variation of slang between the US and the UK,  two  prominent English-speaking communities where data is readily available, even though our computational framework can be applied generally to other communities. Ideally, we would like to extend our work to study more fine-grained communities. This could include, for example, English-speaking regions of smaller size within a country, online communities such as Reddit and Discord, and communities that speak languages other than English.

Another open question is whether the frequency of a slang sense plays a role in determining slang semantic variation and if so in what ways. Our work is limited in this respect since most slang senses in published dictionaries are those that have been prominent enough to be recorded in the first place, and many of these slang usages may have been conventionalized. It would be interesting to study niche slang such as those in the Urban Dictionary if reliable annotations become available.

\section*{Ethics Statement}

We acknowledge that many slang entries analyzed in our work might contain offensive or stereotypical views. We strive to only present examples and illustrations that do not contain such information, by not showing senses that are offensive or stereotypical in nature and editing definition and example usage sentences from the original source to remove potentially offensive content. Discretion is advised when consulting the original data source.

We have obtained written permissions from the author of Green's Dictionary of Slang to use it for personal research use.
We therefore cannot share the full dataset with the public but provide detailed preprocessing scripts and the necessary word lists to facilitate reproduction of our work.

\section*{Acknowledgements}
	
We thank the anonymous EMNLP reviewers for their constructive comments and suggestions. We thank Jonathon Green for permission to use The Green's Dictionary of Slang. YX was supported by an Amazon Research Award, a NSERC Discovery Grant RGPIN-2018-05872, a SSHRC Insight Grant \#435190272, and an Ontario ERA Award.


\bibliography{emnlp2022}
\bibliographystyle{acl_natbib}

\newpage
\appendix

\section{Wiktionary Experiment} \label{appwiki}

Results in Figure~\ref{figwiki} are obtained by counting the number of sense entries in the English Wiktionary.\footnote{\url{https://en.wiktionary.org/}} We extract all word entries using WikExtract~\cite{ylonen22} and only consider those that have 1) at least one slang sense and 2) senses in both US and UK. We determine whether a sense is regional and/or slang using metadata tags associated with each sense. Table~\ref{tablewikitags} shows the full list of tags used. If one of the tags is found in the metadata, then the entry is considered part of the corresponding category. Entries with both US and UK tags are considered neither US specific nor UK specific. We obtain 810 slang words after filtering using the criteria above which contain 8,769 conventional senses and 1,262 slang senses. The proportion of senses with regional tags is shown in Figure~\ref{figwiki}.

\section{Australian Slang} \label{appaus}

We evaluate all semantic distinction models on a 3-way classification task including Australian slang, which makes up a small but notable portion of GDoS. Here, we only consider words with at least 3 regional senses (i.e., $k=3$) due to data sparsity. Also, communivative need models are not evaluated because frequency statistics are not available for the Australian region on Google Ngram. We find 44 word entries that match the critera with 395, 254, and 167 regional entries for US, UK, and Australia respectively. We also include 467 shared senses similar to the experiment described in Section~\ref{secmainexp}. We  sample class-balanced test sets and obtain 309 examples evenly divided among the 3 regions. We repeat this sampling procedure 20 times to account for all senses. Table~\ref{tableaus} shows the results. We observe similar trends as in the US and UK only case where the semantic chaining models substantially outperform all baselines. The exemplar model achieves the highest predictive accuracy both overall and on the Australian test cases despite the set of Australian historical senses being less frequent than the others.

\section{Sample Sizes for Historical Analysis} \label{apphistc}

Table~\ref{tablehistk} shows the number of entries sampled for the class-balanced test samples used in Section~\ref{sechist}. Each sample contains an even number of entries from US and UK.

\section{Hybrid Model Analysis} \label{appcombo}

We consider a hybrid model involving both communicative need and semantic distinction by combining their likelihood functions:

\begin{align}
    \mathcal{V}(s, w, r) \propto \mathcal{V}_{\text{need}}(\cdot) * \mathcal{V}_{\text{distinction}}(\cdot)
\end{align}
Table~\ref{tablecombo} shows an extended version of Table~\ref{tablemain} with the hybrid models included. We observe that the joint model often performs much worse than the  individual models, largely due to discrepancies between the two models predictions. This result has been partly reflected in the analysis presented in Section~\ref{sechist} and Figure~\ref{figreshist}. There, we observe that the relative performance of each hypothesis-driven model to vary substantially point-by-point over the time course. Therefore, combining both models often results in degraded overall predictability  (e.g., often a good model is paired with an impoverished model at a certain time).

\begin{table*}[t!]
	\centering\makebox[\textwidth]{
		\begin{tabular}{ll}
		    
			Category&Tags
			\\
			\addlinespace[0.05cm]
			\hline
			
			\addlinespace[0.1cm]
			Slang & Cockney, informal, slang, vulgar\\
			\addlinespace[0.05cm]
			\makecell[l]{US\\\\\\} & \makecell[l]{Boston, California, Florida, Louisiana, Maine, Midwestern-US, New-Jersey, New-York, \\New-York-City, North-America, Northern-US, Pennsylvania, Philadelphia, Southern-US,\\ Texas, US, Virginia, in US, in US and Canada, in US usually formal, in the US}\\
			\addlinespace[0.05cm]
			\makecell[l]{UK\\\\\\\\\\} & \makecell[l]{Britain, British, Cornwall, Derbyshire, Devon, East-Anglia, England, Kent, Liverpudlian, \\Mackem, Midlands, Multicultural-London-English, Norfolk, Northern-England, \\Northern-English, Northumbria, Orkney, Oxford, Pembrokeshire, Scotland, Shetland, \\Teesside, Tyneside, UK, Ulster, Wales, West-Midlands, Yorkshire, in Britain, in UK, \\of England}\\

	\end{tabular}}
	\caption{Wiktionary metadata tags used to determine whether a sense is a slang or belongs to US or UK.}
	\label{tablewikitags}
\end{table*}

\begin{table*}[t!]
	\centering\makebox[\textwidth]{
		\begin{tabular}{lrrrr}
		    
			Model&US senses&UK senses&AUS senses&All senses
			\\
			\addlinespace[0.05cm]
			\hline
			
			\addlinespace[0.10cm]
			[No shared senses] & \\
			
			\addlinespace[0.1cm]
			Sense frequency & 58.4 (2.58) & 25.3 (1.95) & 4.8 (0.78) & 29.5 (1.01)\\
			LDA & 58.1 (3.56) & 23.4 (1.42) & 11.8 (1.29) & 31.1 (1.38)\\
			Logistic reg. & 54.3 (3.33) & 25.5 (1.74) & 10.1 (1.08) & 30.0 (1.34)\\
			
			\addlinespace[0.15cm]
			Onenn & 52.4 (2.80) & 26.2 (1.75) & 28.3 (1.52) & 35.6 (1.15)\\
			Exemplar & 42.1 (3.22) & 30.7 (1.52) & 38.9 (1.68) & \textbf{37.2} (1.34)\\
			Prototype & 50.4 (3.26) & 29.4 (1.71) & 22.4 (1.64) & 34.0 (1.34)\\
			
			\addlinespace[0.10cm]
			[With shared senses] & \\
			
			\addlinespace[0.10cm]
			Sense frequency & 54.3 (2.85) & 31.3 (1.58) & 1.2 (0.42) & 28.9 (1.11)\\
			LDA & 40.5 (3.28) & 23.4 (1.42) & 11.8 (1.29) & 25.3 (1.21)\\
			Logistic reg. & 56.5 (2.97) & 26.5 (1.74) & 7.8 (1.11) & 30.2 (1.17)\\
			
			\addlinespace[0.15cm]
			Onenn & 67.9 (2.74) & 23.1 (2.17) & 23.0 (1.03) & 38.0 (1.21)\\
			Exemplar & 45.1 (3.06) & 27.0 (1.49) & 49.3 (1.70) & \textbf{40.5} (1.32)\\
			Prototype & 51.7 (3.50) & 32.1 (1.95) & 31.7 (1.20) & 38.5 (1.56)\\
	\end{tabular}}
	\caption{Mean percentage accuracy of all models on the region tracing task for post 1900 senses associated with words that have at least 3 regional senses in each region (US, UK and AUS; $k=3$). Standard deviation of accuracies taken across 20 test samples is shown in parenthesis.}
	\label{tableaus}
\end{table*}

\begin{table}[t!]
	\centering\makebox[0.5\textwidth]{
		\begin{tabular}{lrrrrr}
			Decade&Test set sample size
			\\
			\addlinespace[0.05cm]
			\hline
			\addlinespace[0.1cm]
			1900 &104\\
			\addlinespace[0.05cm]
			1910 &48\\
			\addlinespace[0.05cm]
			1920 &44\\
			\addlinespace[0.05cm]
			1930 &92\\
			\addlinespace[0.05cm]
			1940 &20\\
			\addlinespace[0.05cm]
			1950 &62\\
			\addlinespace[0.05cm]
			1960 &64\\
			\addlinespace[0.05cm]
			1970 &74\\
			\addlinespace[0.05cm]
			1980 &94\\
			\addlinespace[0.05cm]
			1990 &104\\
			\addlinespace[0.05cm]
			2000 &92\\
			\addlinespace[0.05cm]
			2010 &26\\
	\end{tabular}}
	\caption{Number of GDoS sense entries sampled for each time period (in decades). A sense's time of emergence is determined by the time tag associated with the sense entry's first valid reference.}
	\label{tablehistk}
\end{table}

\begin{table*}[t!]
	\centering\makebox[\textwidth]{
		\begin{tabular}{lrrrrrr}
		    
		    &\multicolumn{3}{c}{No shared senses}&\multicolumn{3}{c}{With shared senses}
			\\
			Model&US senses&UK senses&All senses&US senses&UK senses&All senses
			\\
			\addlinespace[0.05cm]
			\hline
			
			\addlinespace[0.1cm]
			Form need & 49.9 (1.20) & 50.7 (0.20) & 50.3 (0.57)\\
			Semantic need & 54.0 (1.77) & 50.1 (0.43) & 52.1 (0.91) & \multicolumn{3}{c}{Not applicable}\\
			Context need & 65.9 (1.37) & 43.8 (0.46) & \textbf{54.8} (0.67)\\

			\addlinespace[0.15cm]
			Sense frequency & 55.9 (1.44) & 34.1 (0.26) & 45.0 (0.75) & 59.2 (1.33) & 34.0 (0.27) & 46.6 (0.66)\\
			LDA & 51.7 (1.37) & 45.3 (0.35) & 48.5 (0.73) & 54.7 (1.46) & 45.5 (0.50) & 50.1 (0.81)\\
			Logistic reg. & 52.5 (1.57) & 40.0 (0.35) & 46.2 (0.84) & 56.5 (1.24) & 39.3 (0.32) & 47.9 (0.64)\\
			
			\addlinespace[0.15cm]
			Onenn & 60.9 (1.35) & 53.0 (0.42) & 56.9 (0.71) & 72.4 (1.32) & 38.0 (0.41) & 55.2 (0.64)\\
            \addlinespace[0.05cm]
            \hspace{0.5cm} + Form & 52.9 (1.33) & 49.0 (0.55) & 50.9 (0.59) & 53.0 (1.43) & 46.0 (0.37) & 49.5 (0.64)\\
            \hspace{0.5cm} + Semantic & 51.1 (1.52) & 55.9 (0.45) & 53.5 (0.71) & 52.6 (1.55) & 53.9 (0.40) & 53.2 (0.76)\\
            \hspace{0.5cm} + Context & 55.0 (1.45) & 61.1 (0.50) & 58.0 (0.65) & 56.3 (1.43) & 53.5 (0.34) & 54.9 (0.72)\\
            \addlinespace[0.15cm]
            
			Exemplar & 60.1 (1.31) & 57.8 (0.40) & \textbf{58.9} (0.70) & 60.0 (1.66) & 58.6 (0.38) & \textbf{59.3} (0.85)\\
            \addlinespace[0.05cm]
            \hspace{0.5cm} + Form & 59.8 (1.25) & 53.0 (0.40) & 56.4 (0.66) & 52.3 (1.61) & 50.3 (0.39) & 51.3 (0.76)\\
            \hspace{0.5cm} + Semantic & 59.8 (1.24) & 53.0 (0.37) & 56.4 (0.67) & 50.0 (1.39) & 57.0 (0.41) & 53.5 (0.72)\\
            \hspace{0.5cm} + Context & 58.7 (1.34) & 54.1 (0.39) & 56.4 (0.70) & 50.8 (1.55) & 65.3 (0.34) & 58.1 (0.75)\\
            \addlinespace[0.15cm]
            
			Prototype & 57.6 (1.38) & 53.1 (0.37) & 55.4 (0.77) & 60.3 (1.71) & 54.7 (0.30) & 57.5 (0.88)\\
            \addlinespace[0.05cm]
            \hspace{0.5cm} + Form & 49.2 (1.54) & 49.4 (0.40) & 49.3 (0.74) & 52.0 (1.75) & 50.0 (0.45) & 51.0 (0.90)\\
            \hspace{0.5cm} + Semantic & 47.7 (1.45) & 56.3 (0.43) & 52.0 (0.69) & 51.4 (1.83) & 55.1 (0.45) & 53.2 (0.82)\\
            \hspace{0.5cm} + Context & 48.9 (1.42) & 60.6 (0.44) & 54.7 (0.61) & 50.0 (1.84) & 61.9 (0.37) & 56.0 (0.82)\\
	\end{tabular}}
	\caption{Mean percentage accuracy of all models (including hybrid models combining both communicative need and semantic distinction) on the region tracing task for post 1900 senses associated with words that have at least 5 regional senses in each region (US and UK; $k=5$). Standard deviation of accuracies taken across 20 test samples is shown in parenthesis. The right-hand side shows the results after including shared senses in both training and prediction.}
	\label{tablecombo}
\end{table*}

\end{document}